%% file: typenet.tex
\documentclass{article}

\usepackage[final]{nips_2017}

\usepackage[utf8]{inputenc} % allow utf-8 input
\usepackage[T1]{fontenc}    % use 8-bit T1 fonts
\usepackage{hyperref}       % hyperlinks
\usepackage{url}            % simple URL typesetting
\usepackage{booktabs}       % professional-quality tables
\usepackage{amsfonts}       % blackboard math symbols
\usepackage{nicefrac}       % compact symbols for 1/2, etc.
\usepackage{microtype}      % microtypography
\usepackage{xcolor}
\usepackage{url}
\usepackage{wrapfig}
\usepackage{latexsym}
\usepackage{multirow}
\usepackage{mathtools}
\usepackage{calc}
\usepackage{amsmath, amsthm, amssymb, amsfonts}
\PassOptionsToPackage{numbers, compress}{natbib}

\newcommand\norm[1]{\left\lVert#1\right\rVert}

\usepackage{graphicx}
\usepackage[export]{adjustbox}

\DeclareMathOperator{\ReLU}{ReLU}

\title{Finer Grained Entity Typing with TypeNet}

\author{Shikhar Murty, Patrick Verga, Luke Vilnis, Andrew McCallum \\
College of Information and Computer Sciences\\
University of Massachusetts Amherst\\
\texttt{\{smurty, pat, luke, mccallum\}@cs.umass.edu}}

\begin{document}

\maketitle
\begin{abstract}
We consider the challenging problem of entity typing over an extremely fine grained set of types, wherein a single mention or entity can have many simultaneous and often hierarchically-structured types. Despite the importance of the problem, there is a relative lack of resources in the form of fine-grained, deep type hierarchies aligned to existing knowledge bases. In response, we introduce TypeNet, a dataset of entity types consisting of over 1941 types organized in a hierarchy, obtained by manually annotating a mapping from 1081 Freebase types to WordNet. We also experiment with several models comparable to state-of-the-art systems and explore techniques to incorporate a structure loss on the hierarchy with the standard mention typing loss, as a first step towards future research on this dataset.
\end{abstract}

\input{intro}

\input{dataset}

\input{model}

\input{results}

\input{related-work}

\section{Conclusion and Future Work}
We introduced TypeNet, a human labeled alignment between Freebase entity types and WordNet synsets. We used this typeset to distantly label the CoNLL-YAGO entity linking dataset and reported initial results with several models comparable to the state-of-the-art models previously used on pre-existing datasets e.g. \citet{shimaoka-EtAl:2017:EACLlong}. 

% We additionally present results from models incorporating a structure loss over the type hierarchy, which does not appear to be required by the CoNLL-YAGO dataset, which is mostly annotated with leaf types, precluding situations where a non-structured model would incorrectly predict invalid child types. 

We additionally present results from models incorporating a structure loss over the type hierarchy, which does not appear to be required by the CoNLL-YAGO dataset, but should be helpful on more diverse datasets from different domains e.g. ClueWeb, which we will explore in future work and encourage the community to do the same.

We are exploring more sophisticated methods of incorporating the type hierarchy into the typing loss, as well as joint models for related tasks such as simultaneous typing and entity linking.

We are excited to see what the community will do with TypeNet, the largest and deepest entity type hierarchy with manual alignment to Freebase. We hope this will spur improvements in fine-grained entity and mention typing, linking and associated downstream tasks.

%\clearpage
%\newpage
\bibliography{sources}
\bibliographystyle{emnlp_natbib}

\end{document}

%% file: intro.tex
\section{Introduction}
\label{sec:intro}
% Recognizing entities and their types is an essential task in language understanding and necessary for down stream tasks such as relation extraction, knowledge base construction, question answering, and query understanding. Early attempts at entity recognition focused only on very coarse grained types (people, locations, organizations) \citep{tjong2003introduction,hovy2006ontonotes}. More recently, there has been growing interest in finer grained entity typing \citep{ling2012fine}.

Recognizing entities and their types is a core problem in natural language processing, underlying complex natural language understanding problems in relation extraction \citep{yaghoobzadeh-adel-schutze:2017:EACLlong}, knowledge base construction, question answering \citep{lee2006fine}, and query comprehension \citep{dalton2014entity}. Early attempts at entity recognition focused only on very coarse grained types \citep{tjong2003introduction,hovy2006ontonotes}. More recently, there has been growing interest in models explicitly focused on entity typing with finer grained typesets e.g. FIGER \citep{ling2012fine}.

The increasingly sophisticated natural language understanding tasks undertaken by the machine learning community often require commensurately more sophisticated world knowledge. This world knowledge, often organized hierarchically in \emph{ontologies}, motivates our creation of a new fine-grained, deep, and high-quality dataset of hierarchical types.

% \todo{motivate hierarchical types and need for new dataset. Ours is the first work to explore incorporating hierarchy into the loss objective}

Despite the increasing focus on fine-grained typing, existing typesets still contain only on the order of 100 different types. Further, these typesets are either endowed with only a shallow hierarchy, typically on the order of two levels deep or don't have links to existing KBs (see Table \ref{data:stats}). In this work, we advocate for larger, deeper typesets and models that exploit the inherently hierarchical nature of these types. To this end, we present TypeNet, an expert-annotated type hierarchy containing 1941 individual types, with an average depth of 7.8. 

%built on top of 1110 freebase\citep{bollacker2008freebase} types which are then mapped onto the WordNet hierarchy \cite{fellbaum1998wordnet}. This gives us a set of over 2132 types, 20x times bigger than FIGER, arranged in a hierarchy.

We also evaluate several models for fine-grained entity typing, and establish a strong baseline of 74.8 MAP on the CoNLL-YAGO dataset \citep{hoffart2011robust} for our best model. With each entity having on the order of 30 types, there are clearly exciting opportunities for improvement from future research. Additionally, we investigate multi-task models that explicitly incorporate the hierarchical relations between types into the learning objective.

%% file: dataset.tex
\section{Dataset Creation}
\label{sec:dataset}
\begin{figure}
%   \begin{minipage}[c]{0.45\textwidth}
    \hfill{\includegraphics[scale=1.08,left]{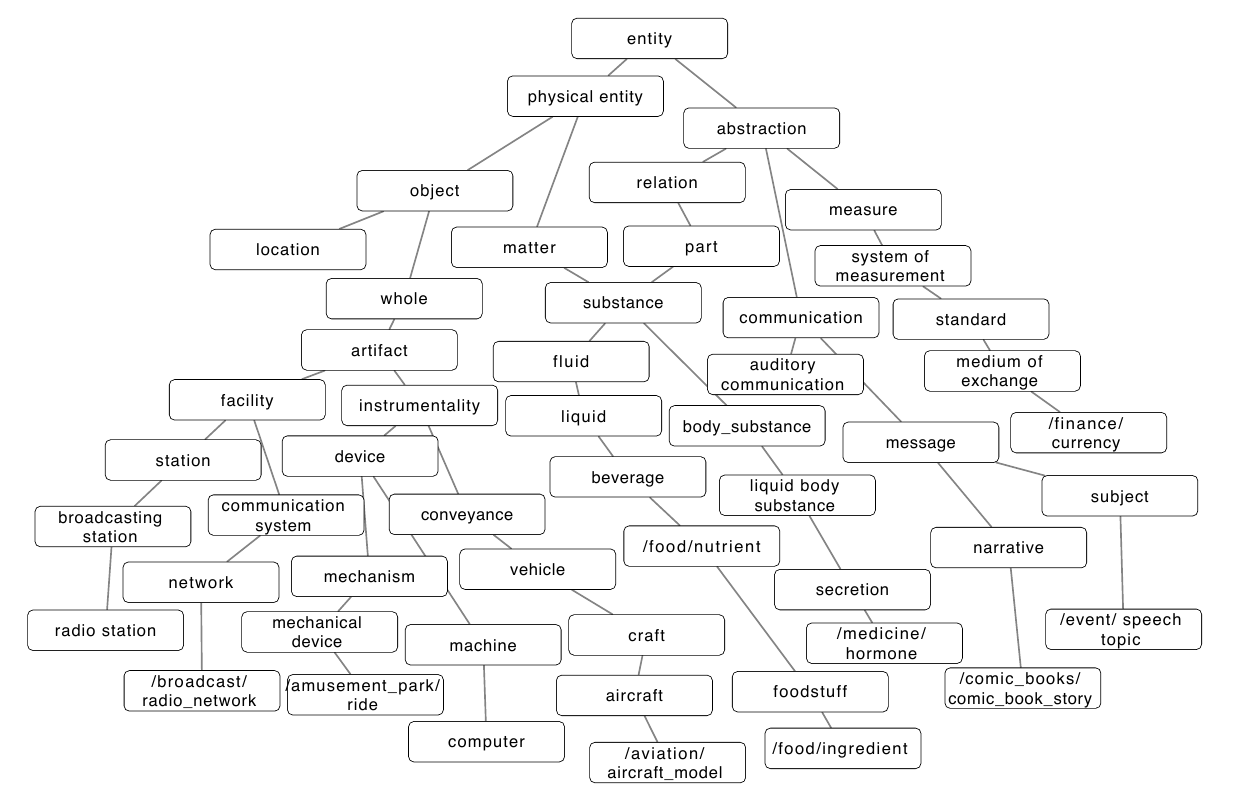}}
%   \end{minipage}\hfill
%   \begin{minipage}[c]{0.35\textwidth}
    \caption{Sampled types in the TypeNet hierarchy \label{fig:network}}
%   \end{minipage}
\end{figure}

We now discuss TypeNet\footnote{https://github.com/iesl/TypeNet}, a new dataset of entity types for extremely fine grained entity typing. TypeNet was created by manually aligning Freebase types to noun synsets from the WordNet hierarchy \citep{fellbaum1998wordnet}, naturally producing a hierarchical type set.

This was done by first filtering out all Freebase types that were linked to $\geq$ 20 entities, and then filtering Freebase API types. The Freebase API types we filtered were those in the domain "/freebase","/dataworld","/schema", "/atom", "/scheme" and "/topics". 

For each Freebase type in our filtered set, we generate a list of candidate WordNet synsets through a substring match. The annotator then attempted to map the Freebase type to one or more synset in the candidate list with a \emph{parent-of}, \emph{child-of} or \emph{equivalence} link by examining definitions of each synset and example entities of the Freebase type. If no match was found, the annotator queried the online WordNet API until an appropriate synset was found.

This procedure was carried out by two separate annotators independently after which conflicts were discussed and resolved. The annotators were conservative with assigning equivalence links resulting in a greater number of \emph{child-of} links. The final dataset contained 13 \emph{parent-of}, 727 \emph{child-of}, and 380 \emph{equivalence} links. Note that some Freebase types have multiple \emph{child-of} links to WordNet. Finally, all the ancestors of the Freeebase types (following the manually created \emph{child-of} and WordNet hypernym links) were added to construct the dataset.

We also carried out a procedure to add an additional set of 614 \emph{fb} $\rightarrow$ \emph{fb} links. This was done by computing conditional probabilities of freebase types given other freebase types from a collection of 5 million randomly chosen freebase entities. We then threshold these probabilities to 0.7, and manually filter the resulting links. 

%We manually mapped our set of 1110 freebase entity types as equivalent to, parent of, or child of one or more synsets. Each mapping was agreed upon by two human annotators. 
\label{sec:data_stats}
\begin{table}[htp]
\centering
\begin{tabular}{llcc}
    Typeset & Count & Depth & Gold KB links \\ \hline \hline
    CoNLL-YAGO &  4  & 1 & Yes \\
    Ontonotes &  19  & 1 & No \\
    \citet{DBLP:journals/corr/GillickLGKH14} & 93 & 3 & Yes \\
    Figer & 121 & 2 & Yes \\
    Hyena & 505 & 9 & No\\
    Freebase & 2k & 2 & Yes \\
	WordNet & 16k & 14 & No \\ 
    TypeNet & 1941* & 14 & Yes \\
  \end{tabular}
  \caption{Statistics from various type sets. TypeNet is the largest type hierarchy with a gold mapping to KB entities. *The entire WordNet could be added to TypeNet increasing the total size to 17k types.  \label{data:stats}}
\end{table}

%% file: model.tex
\section{Model}
\label{sec:model}
We now describe the various neural models we use for our experiments.

\begin{figure}
%   \begin{minipage}[c]{0.45\textwidth}
    \hfill{\includegraphics[scale=.6,left]{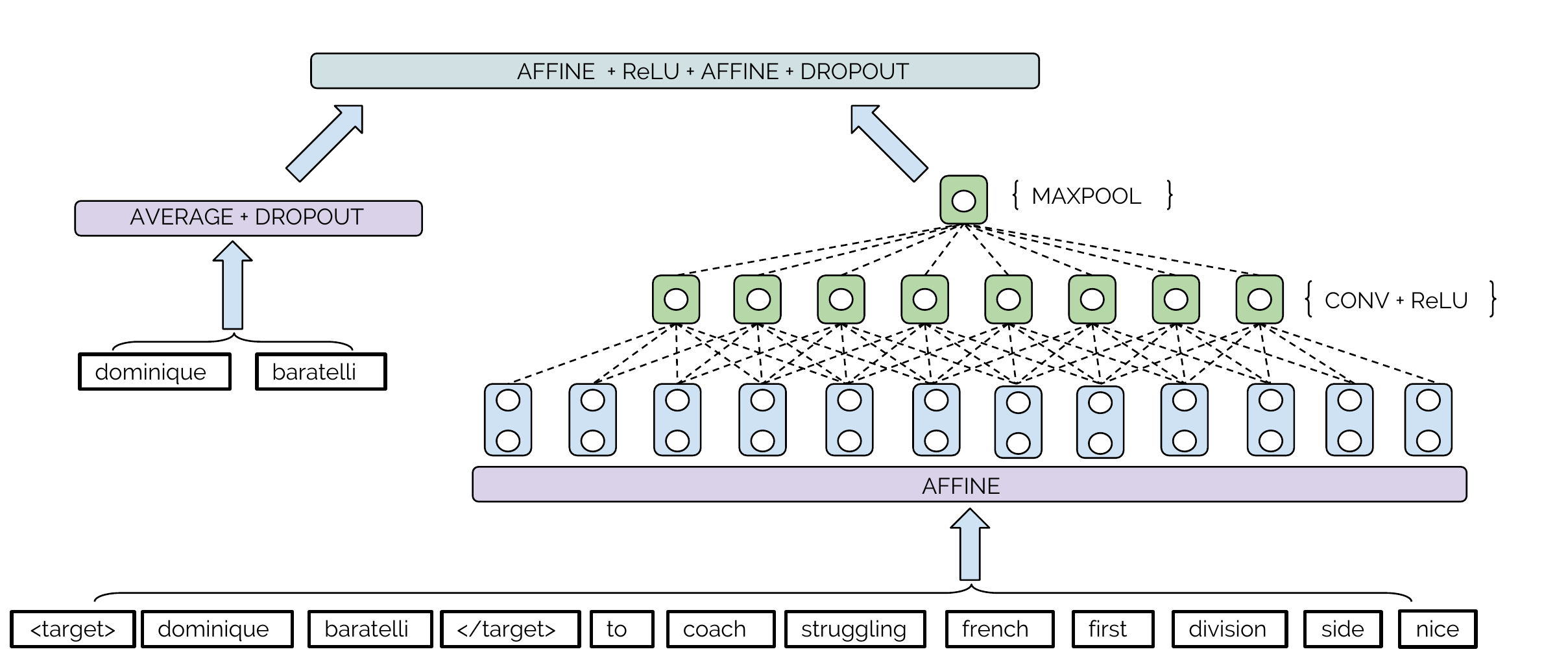}}
%   \end{minipage}\hfill
%   \begin{minipage}[c]{0.35\textwidth}
    \caption{Mention Encoder} 
    
    %The mention sentence is encoded as follows. We run a convolution with a filter of width 5 over word embeddings of the mention tokens, zero padded to the right. The token embeddings are obtained by passing pre-trained embeddings through an affine transform, followed by a ReLU operation \citep{glorot2011deep}. The outputs of the convolutions are then max pooled to give a single vector, which is passed through a dropout layer. This representation is then concatenated with another vector representating the average of the surface form tokens of the entity linked in this mention, and the concatenation is passed through another affine layer followed by a ReLU, an affine layer and a dropout layer to get the final mention representation. 
%   \end{minipage}
\label{fig:encoder_arch}
\end{figure}

\noindent{\bf Input Layer:} We represent a mention $m = (\vec{m_1}, \vec{m_2},...,\vec{m_n})$ as a sequence of word vectors where each vector is of a fixed dimension $d$. To obtain a mention vector representation, we use a Convolutional Neural Network (CNN) based architecture. The CNN learns mention representations from sliding w-gram features of the mention. For a mention $m$ with $n$ tokens represented as vectors, the CNN outputs $n-w+1$ vectors, which are then max-pooled along every dimension to obtain $\vec{m}_{\text{CNN}}$:
\begin{align}
\vec{x_j} &= \ReLU(\vec{b} + \sum_{k=0}^{w} W[k:] \cdot \vec{m}_{[j-{\frac{w}{2}}+k]}) \\
\vec{m}_{\text{CNN}} &= \max\{\vec{x}_1, \vec{x}_2, ..., \vec{x}_{n-w+1}\} 
\end{align}

where $W$ is a CNN filter of size $w \times d \times d$, and $\vec{b}$ is a bias vector of size $d$.  We then concatenate $\vec{m}_{\text{CNN}}$ with another vector $\vec{m}_{\text{SFM}}$ obtained by averaging the surface form of the entity to which the mention links. We do this to provide our model  explicit signal about the entity present in the mention. The concatenation is then passed through a series of affine, ReLU and affine transforms to obtain the final mention representation $\vec{\mathsf{m}}$ (See Fig-\ref{fig:encoder_arch}):
\begin{align}
\vec{m}_{\text{SFM}} &= \frac{1}{\vert t_2 - t_1 + 1 \vert} \sum_{k = t_1}^{k= t_2} \vec{x_k} \\
% \vec{\mathsf{m}} &= W_2(\ReLU(W_1 \begin{bmatrix}\vec{m}_{SFM}\end{bmatrix}
\vec{\mathsf{m}} &= W_2(\ReLU(W_1 \begin{bmatrix}\vec{m}_{\text{SFM}} \\ \vec{m}_{\text{CNN}}\end{bmatrix} + b_1) + b_2)
\end{align}

%The mention encoder is also summarized in Fig-\ref{fig:encoder_arch}

\noindent{\bf Loss Function:} Like prior work \citep{shimaoka-EtAl:2017:EACLlong}, we model entity typing as a multi-label problem, and for a given mention, produce a vector of scores corresponding to each type. We optimize a mention-typing loss over each minibatch $B_1$ of $(m,\{t_i\})$ pairs, where $\{t_i\}$ is the set of gold types annotated for the mention $m$:
\begin{align}
\mathcal{L}_{\text{typing}} &= \frac{1}{|B_1|} \sum_{(m,\{t_i\}) \in B_1} \bigg( \sum_{t \in \{t_i\}} -\text{score}(m \in t) ~~~- \sum_{t' \notin \{t_i\}} \text{score}(m \notin t') \bigg)
\end{align}
where $\text{score}(m \in t)$ is some function indicating the compatibility score for mention $m$ being in type $t$, using the interpretation of a type as a set defined by a unary predicate (``has type t'').

In this work, we also introduce a structure loss among the types to incorporate the hierarchy. For this, we have a separate minibatch $B_2$ of $(t,\{t_a\})$ pairs, where $\{t_a\}$ is the set of ancestor types for the type $t$:
\begin{align}
\mathcal{L}_{\text{structure}} &= \frac{1}{|B_2|} \sum_{(t,\{t_i\}) \in B_2} \bigg( \sum_{t_a \in \{t_a\}} -\text{score}(t \in t_a) ~~~- \sum_{t' \notin \{t_a\}} \text{score}(t \notin t') \bigg)
\end{align}
The exact scoring functions used for different models are summarized in Table \ref{table:score-functions}, but are either variations of binary cross entropy or order embedding loss. We experiment with models whose loss is either $\mathcal{L}_{\text{typing}}$ alone, or a weighted combination of $\mathcal{L}_{\text{typing}}$ and $\mathcal{L}_{\text{structure}}$.

\begin{table}[htp]
\centering
\begin{tabular}{l|l|l}
    Model & $\text{score}(x \in y)$ & $\text{score}(x \notin y)$ \\ \hline \hline
    Order Embedding &  -$\norm{\text{max}(0, y-x)} ^2$ &  $\max(0, \alpha-\norm{\text{max}(0, y-x)} ^2)$\\
    Bilinear & ~$\log\sigma (x^{\top}Ay)$ & ~$\log (1 - \sigma (x^{\top}Ay))$\\
    Dot &  ~$\log\sigma (x^{\top}y)$  & ~$\log(1-\sigma (x^{\top}y))$\\
  \end{tabular}
  \vspace{.3cm}
  \caption{Scoring functions used for modeling type membership, where $\sigma$ is the logistic sigmoid, and $A$ is a learned parameter matrix (bilinear form) that scores the asymmetric membership relation.}
  \label{table:score-functions}
\end{table}

% \begin{table}[htp]
% \centering
% \begin{tabular}{l|l}
%     Model & scoring function \\ \hline \hline
%     Order Embedding &  -$\norm{\text{max}(0, y-x)} ^2$ \\
%     Bilinear & $x^{\top}Ay$ \\
%     Dot &  $x^{\top}y$  \\
%   \end{tabular}
%   \caption{Scoring functions used for modelling edges in $\mathcal{T}$ \label{model:equations}}
% \end{table}

\noindent{\bf Hyperparameters:} We use pretrained 300 dimensional case sensitive GloVe vectors by \citet{DBLP:conf/emnlp/PenningtonSM14} and a CNN with a filter width of 5. The type vectors are all 300 dimensional initialized using Glorot initialization \citep{glorot2011deep}. We use dropout \citep{DBLP:journals/jmlr/SrivastavaHKSS14} as described in Fig-\ref{fig:encoder_arch} and optimize using Adam \citep{DBLP:journals/corr/KingmaB14}. We tune our hyperparameters via grid search and early stopping on the development set.

%% file: results.tex
\section{Results}
\label{sec:results}

\noindent{\bf Dataset and Evaluation metrics:} To perform our experiments, we use the CoNLL-YAGO \citep{hoffart2011robust} dev/test split and a subsampled version of Wikipedia (2016/09/20 dump) for training. We obtain labels for mentions via \emph{distant supervision}, by assuming as positive types \emph{all} the TypeNet types of the entity linked to a mention. We do not perform any heuristic pruning/denoising of these types, even though not all of them are relevant for a mention. This is because most pruning methods in literature are either harsh for extremely fine types \citep{DBLP:journals/corr/GillickLGKH14}, or did not give an increase in performance \citep{shimaoka-EtAl:2017:EACLlong}. However, we plan on improving this in future work and release a gold test data set.

To obtain the TypeNet types of an entity, we filter out all its Freebase types present in TypeNet, and finally for every type filtered out, add all its ancestors from TypeNet, giving us an average of 30.73 types per mention, \emph{much} greater than earlier datasets such as FIGER (GOLD) \citep{ling2012fine} for which the average was 1.73 types per mention. 

Since we have on average 30x more types per entities, we use Mean Average Precision (MAP) to measure performance unlike prior work on fine grained entity typing \citep{shimaoka-EtAl:2017:EACLlong}. Our results are summarized in Table \ref{tab:typenet}. 

\noindent{\bf Discussion:} We observe that the CNN encoder model works best, and multitasking mention typing with structure decreases performance if the structure is modeled using a dot product. This is expected since hypernymy is an asymmetric relation. Furthermore, modeling structure with a bilinear objective improves performance over a dot product objective but fails to perform better than the regular model. We believe this is due to the nature of our test data which is made up mostly of leaf type predictions from a non-diverse set of types (people, locations, and organizations). A more diverse dataset requiring predictions at different depths of the hierarchy could use this structure more effectively, since we posit that without using structure, the model will incorrectly predict leaf types when only parent types are true.

Interestingly, we observe the order embeddings model from \cite{vendrov2015order} to have a poor performance for our task. We attribute this to the fact that the loss function is poorly suited to the problem since it uses unrelated concepts as negative examples, which in the traditional order embedding model actually implies a reversal of the parent-child relation, rather than simply forcing the types to be unrelated. For example, consider a hypernym link \emph{person} $\Rightarrow$ \emph{organism}, and a negative example, \emph{person} $\Rightarrow$ \emph{stadium}. The loss function from \cite{vendrov2015order} attempts to increase the order violation between \emph{person} and \emph{stadium}, making \emph{stadium} a hyponym of \emph{person}. We also observe particularly poor performance combining the order embeddings with the CNN encoder.

\begin{table}
\centering
\begin{tabular}{c|l|l}
    \multicolumn{2}{c|}{Model} & MAP  \\ \hline \hline
   \multirow{4}{*}{Order}	  	  & Mention  &  50.2 \\
                                  & +structure & 50.7  \\
                                  \cline{2-3}
                                  & CNN & 44.0 \\
                                  & +structure & 44.6 \\
   \hline
    \multirow{5}{*}{Bilinear}     & Mention  &  70.0 \\
    							  & +structure (dot) &  69.5 \\
                                  \cline{2-3}
    							  & CNN  & \textbf{74.8}  \\
    						  	  & +structure (dot) & 73.5 \\
                                  & +structure (bilinear) &  \textbf{74.8} \\
    \hline
  \end{tabular}
  \caption{Mean average precision for various models on TypeNet. "Mention" refers to simply averaging the words of the entity mention surface form. CNN concatenates the mention representations with a sentence representation. Scores with +structure additionally multi-task the structure loss training objective. \label{tab:typenet}}
\end{table}

%% file: related-work.tex
\section{Related work}

Table-\ref{data:stats} summarizes existing hierarchical type systems including popular data sets such as FIGER \citep{ling2012fine} and \citet{DBLP:journals/corr/GillickLGKH14}.

\citet{delcorro-EtAl:2015:EMNLP} considered the task of extremely fine grained entity typing.
They use manually crafted rules and patterns (Hearst patters, appositives, etc). \citet{hearst1992automatic} to extract candidate entity types that match Wordnet synsets. They apply an optional KB type filtering step for entity $e$ by matching a candidate type $t_c$ to any of $e$'s KB types $T$ if $t_c$ is a string match of any $t_i$, $hypernym(t_i)$, or $hyponym(t_i)$. We instead manually annotated the exact mapping from 1081 Freebase types to the specific WordNet synset sense allowing us to leverage distant supervision to trained supervised classifiers.

The knowledge base Yago \citep{DBLP:journals/ws/SuchanekKW08} includes integration with WordNet and type hierarchies have been derived from its type system \citep{yosef2012hyena}. However the links between entity types and WordNet types are performed heuristically whereas TypeNet contains gold links between Freebase and WordNet.

There has been a growing interest in learning representations of hierarchically organized objects. \citet{vilnis2014word} proposed Gaussian embeddings which learn containment properties of words by approximating them with Gaussian distributions. \citet{vendrov2015order} introduced order embeddings by minimizing an order violation loss. Recently \citet{nickel2017poincar} proposed Poincar{\' e} embeddings.